# Real Time Estimation of Bayesian Networks

**Robert L. Welch**
**Gensym Corporation**
**4940 Pearl East Circle**
**Boulder, Colorado 80303**

### Abstract

For real time evaluation of a Bayesian network when there is not sufficient time to obtain an exact solution, a guaranteed response time, approximate solution is required. It is shown that non traditional methods utilizing estimators based on an archive of trial solutions and genetic search can provide an approximate solution that is considerably superior to the traditional Monte Carlo simulation methods.

## INTRODUCTION

Large complex Bayesian networks (such as would be required for diagnosing a power plant) are a challenge to solve. Approximation methods are employed when the demands of a real time application require an answer within a time period shorter than is needed for an exact solution. Most approximation methods are based on a Monte Carlo style of simulation: Trial solutions are generated independently and a frequency distribution is accumulated for each node in the network. The relative frequency with which a node is observed to be in a particular state is an approximation to the conditional probability of the node given the values of the instantiated nodes in the network (the evidence).

In real time situations, a solution to the network is computed periodically with new values for the (possibly changing) set of instantiated nodes. This repetitive evaluation of the network allows for non-traditional simulation methods that take advantage of the history of previous evaluations. In this paper we explore two innovations for simulation that can provide an answer of given accuracy considerably faster than traditional Monte Carlo style methods. One algorithm maintains an archive of the unique trials that have been generated. The network solution is computed based on summing the joint probability of trials that are consistent with the evidence.

A second innovation is to utilize a genetic algorithm[1] to search for new trials solutions that have the greatest joint probability within that portion of the solution space conforming to the evidence.

In a large Bayesian network, the prior probability of the evidence is often very small. This is especially true when these values are the symptoms of real world phenomena associated with a faulty condition or accident. These low probability cases are the situations for which Bayesian networks have perhaps the greatest value as an analytical device. Even so, the low probability presents a challenge to Monte Carlo simulation since only a small number of the trials will conform to the evidence and the solution is likely to be more greatly influenced by the evidence than by the prior probability of the root nodes. Recent developments in Monte Carlo based methods for simulating Bayesian networks have sought to place greater weight on evidence rather than on prior probabilities[2]. In computing the frequency distribution, these methods weight each new trial solution by the likelihood that the trial (which comes from a known sampling distribution) could have come from the network's joint probability distribution. Nevertheless, the methods still are rather slow to converge.

The methodology proposed in this paper is based on a view of simulation as a search for the trials solutions that together constitute a large portion of the probability mass of solutions conforming to the evidence. By means of examples which have been selected at random it is shown that a search method such as a genetic algorithm can dramatically improve this search especially when the evidence is of low prior probability and is introduced sequentially as in a real time application.

This paper has the following organization. We first provide a brief description of several Monte Carlo simulation methods for Bayesian networks. Following this introduction, we introduce archive based estimators of the solution. The genetic search algorithm is developed in section 3. Finally we compare the methods with an example (section 4), give evidence as to the generality of the results (section 5) and provide a strategy for real time estimation in the conclusion.

## 1. MONTE CARLO SIMULATION.

---

[1] Goldberg [1989].

[2] Henrion [1986], Chavez & Cooper [1990], Fung & Chang [1989], Schacter & Peot [1989], Fung & Del Favero [1994].

We use the term Monte Carlo simulation to refer to any simulation method in which: 1) Each trial solution is sampled from the same sampling distribution independently and with replacement. 2) The method maintains a running tally of the frequency with which each possible solution is generated. The tally is a weighted sum of the occurrences. 3) The frequencies are normalized to form a relative frequency distribution. The relative frequency is an estimate of the exact solution -- the conditional probability of each node given the evidence (instantiated values of observed nodes in the network).

A Bayesian network is defined by a set of nodes, $N$, a set of directed arcs $V$ connecting pairs of nodes in $N$, and a set of link matrices $P(\alpha \mid \xi(\alpha))$ giving the conditional probability distribution of a node $\alpha$ given the node's parents, $\xi(\alpha)$. A trial solution $X_N$ is an assignment of values to each node in the network $N$. The sample space $X_N$ is the set of all possible trials.

A fundamental property of a Bayesian network representation is that the joint probability distribution over trial solutions is the product of the link matrices:

**Joint Probability Theorem[3] :**

$$P(X_N) = \prod_{\alpha \in N} P(X_\alpha \mid X_{\xi(\alpha)})$$

The network solution is a set of conditional probabilities $P(X_\alpha \mid X_E)$ of each node given the observed values of a selected set of nodes, E. $X_E$ is referred to as the evidence.

In an *ordinary Monte Carlo simulation*, trial solutions are drawn directly from the joint probability distribution $P(.)$ and a tally of the number of times each trial solution is generated during the simulation is maintained. After T trials have been generated, the relative frequency can be computed, $r^T(X_\alpha) = \#\{t \leq T \mid X_\alpha{}^t = X_\alpha\}/ T$. The well known law of large numbers states that for T large, $r^T(X_\alpha)$ is an approximation to $P(X_\alpha)$. In order to incorporate the observed data, $X_E$, the same method could be applied to the posterior joint probability distribution, $P(\cdot \mid X_E)$. However, that distribution is unknown and the discovery of its marginal distributions, $P(X_\alpha \mid X_E)$, is the object of the simulation methods discussed in this paper.

## 1.1 IMPORTANCE SAMPLING[4]

This is a general sampling method for conducting Monte Carlo simulations. It is assumed that the samples (trial solutions) are drawn independently and with replacement from a sampling distribution $P^S(X_N)$. In computing the relative frequency of the generated trials, importance sampling weights each trial by the likelihood of observing the trial given the joint probability $P(\cdot)$.

A simulation method based on importance sampling has the following steps.

- $X_N{}^t$ is drawn independently and at random from the sampling distribution, $P^S(X_N)$.

- A tally $Z^t(\cdot)$ is maintained accumulating a weighted count of the trials generated during simulation: The likelihood that the trial drawn from the sampling distribution could have come from the distribution of interest is $P(X_N{}^t \mid X_E)$ / $P^S(X^t{}_N)$. Since $P(X^t{}_N \mid X_E) = 0$ if $X^t{}_E \neq X_E$ and $= P(X^t{}_N) / P(X_E)$ if $X^t{}_E = X_E$, the weight in the tally is (up to a normalizing constant which can be ignored)

$$z(X^t{}_N) = \begin{cases} 0, \text{ if } X^t{}_E \neq X_E \\ P(X^t{}_N) / P^S(X^t{}_N) \text{ otherwise.} \end{cases}$$

- The tally is incremented as follows: $Z^0(X_N) = 0$;

$$Z^t(X_N) = Z^{t-1}(X_N) + \begin{cases} 0, \text{ if } X_E \neq X_N{}^t \\ \text{else} \\ z(X_N{}^t) . \end{cases}$$

- After the simulation of T trials has completed, normalize the tally $Z^T$ so that it sums to 1 over the sample space. The normalized function is a probability distribution that is an approximation to the joint probability distribution:

$$P^T(X_N) = \kappa Z^T(X_N) \text{ where } \kappa = 1 / \sum_{X_N} Z^T(X_N) \text{ is the normalization factor.}$$

The Importance Sampling Theorem states that $P^T(X_N)$ is indeed an approximation to the joint probability P:

**Importance Sampling Theorem[5].**

$$P^T(X_N) \to P(X_N \mid X_E) \text{ as } T \to \infty.$$

## 1.2 LOGIC SAMPLING SIMULATION[6].

*Logic sampling* is a form of Monte Carlo simulation. The method samples from the joint probability distribution $P()$ ignoring evidence, but all cases that do not agree with the evidence, $X_E$, are discarded. In order to generate a trial from the joint probability it suffices to sample from the link matrices. Arrange the nodes in a precedence ordering $\{\alpha_1 \dots \alpha_N\}$ so that $i < k$ if and only if $\alpha_i$ is a parent of $\alpha_k$, $\alpha_i \in \xi(\alpha_k)$. Then sample each node in order from $P(\alpha \mid \xi(\alpha))$.

---

The resulting approximate distribution is

$$P^T(X_N) = \begin{cases} 0 \text{ if } X_E \neq \underline{X}_E \\ \text{else} \\ \dfrac{\#\{t \leq T | X_N{}^t = X_N \ \& \ X_E{}^t = \underline{X}_E \}}{\#\{t \leq T | X_E{}^t = \underline{X}_E \}} \end{cases}$$

the relative number of times $X_N$ was generated out of the total number of times the evidence was generated.

Although this method does generate an approximate solution that converges to the true solution, convergence is likely to be slow especially when the $P(\underline{X}_E)$ is small. Too much computational effort is expended generating trial solutions that are not consistent with the evidence.

### 1.3 FORWARD SIMULATION[7].

A more efficient simulation method, forward simulation, samples nodes in precedence order. Nodes that are not members of the evidence set E are sampled based on their link matrices. Nodes in the evidence set are given their observed value in $\underline{X}_E$.

The steps of this simulation are:

Generate a value for each node in precedence order from the following sampling distribution:

$$P^S(X_N) = \prod_\alpha P^S(X_\alpha \mid X_{\xi(\alpha)}) \text{ where}$$
$$P^S(X_\alpha \mid X_{\xi(\alpha)}) = 1 \text{ if } \alpha \in E \text{ and } X_\alpha = \underline{X}_\alpha$$
$$= 0 \text{ if } \alpha \in E \text{ and } X_\alpha \neq \underline{X}_\alpha$$
$$= P(X_\alpha \mid X_{\xi(\alpha)}) \text{ otherwise.}$$

$P^S$ forces each node in $E$ to conform with the evidence $\underline{X}_E$.

Then the approximation to the conditional joint probability is

$$P^T(X_N) = 0 \text{ if } X_E \neq \underline{X}_E \text{ else}$$
$$\kappa \#\{t \leq T | X^t{}_N = X_N\} \prod_{\alpha \in E} P(X_\alpha | X_{\xi(\alpha)}).$$

Although this method avoids the wasted computation in Logic Sampling, forward simulation emphasizes the priors of the root nodes. Especially when the observed values have low prior probability, priors are not as important in determining the posterior probabilities as are the observed values.

### 1.4 BACKWARD SIMULATION[8].

This strategy starts from the observed evidence rather than from the root nodes in the network. Starting with the evidence nodes, parent nodes are sampled using Bayes rule to invert the link matrix. This process continues until values for all ancestors of the evidence nodes are generated. Then the remaining nodes are forward sampled using their link matrices. A formal definition is given in the appendix.

## 2. ESTIMATION FROM AN ARCHIVE OF UNIQUE TRIAL SOLUTIONS.

An alternative method depends on maintaining an archive of every unique trial generated during the simulation. The archive is searched for a match with the id code of the new trial. If no match is found, the joint probability of the trial $P(X_N)$ is computed and stored along with the new id code. If there is a match, the new trial is discarded.[9]

The archive is a subset of the sample space $X_N$ that has been revealed through simulation. We are, of course, assuming that $X_N$ is very large, too large to enumerate. In such a large sample space, the joint probability of most of the elements is 0 or very near 0. Simulation methodology can be used to explore the region with most of the probability mass. If the trials generated eventually cover the support of $P(\cdot|\underline{X}_E)$ then alternative estimates of the marginal distributions are obtained by summing the joint probabilities of the trials in the archive. Each time a new trial is added to the archive, we accumulate its joint probability into the sums used to estimate each marginal probability.

Suppose we have an archive of T unique trial solutions $\{X_N{}^1 \dots X_N{}^T\}$. We can compute the following partial sums over trials in the archive:

$P^t = P(X^t{}_N)$ the probability of trial t

$P^T = \sum_{t \leq T} P^t$ the archive probability mass accumulated by T

$P^T(X_A) = \sum_{\{t \leq T | X^t{}_A = X_A\}} P^t$

the archive probability mass conforming to $X_A$

$P^T(\underline{X}_E)$ the archive evidence probability mass conforming to the evidence.

$P^T(X_A \ \& \ \underline{X}_E) = \sum_{\{t \leq T | X^t{}_A = x_A \ \& \ x^t{}_E = x_E\}} P^t$

the archive probability mass conforming to both $X_A$ and $\underline{X}_E$.

The ratio of partial sums, $P^T(X_\alpha \ \& \ \underline{X}_E) \ / \ P^T(\underline{X}_E)$, is the archive estimator of the solution of a Bayesian network. The bias of this estimator depends on the bias in the archive of trial solutions that conform to the evidence. If a process of generating trials solutions can assure that the probability mass of the evidence, $P^T(\underline{X_E})$, grows to eventually cover $P(\underline{X_E})$, then the archive estimator can be used as an approximation to the solution of the Bayesian network.

---

[7] Schacter & Peot [1989].

[8] Fung & Del Favero [1994].

[9] The id code for a trial can be generated from a binary representation of the state assigned to each node in the trial (see figure 3.1).

**Archive Estimation Theorem:** If the process generating trials for the archive can guarantee that $P^T(\underline{X_E}) \to P(\underline{X_E})$ as $T \to \infty$, then for all $\alpha$ in N,

$$P^T(X_\alpha \mid X_E) \equiv P^T(X_\alpha \And \underline{X_E}) \,/\, P^T(\underline{X_E})$$
$$\to P(X_\alpha \mid \underline{X_E}) \text{ as } T \to \infty .$$

Any Monte Carlo simulation will satisfy the condition above and hence can be used as a search process to discover the trials of significant probability that conform to $\underline{X_E}$. When the sample space is finite, the Importance Sampling Theorem assures that every trial that conforms to the evidence $\underline{X_E}$ and has positive probability will eventually be generated.    The frequency estimator traditionally used in Monte Carlo simulation has good convergence properties for a one shot attempt to estimate a Bayesian network. However, when the network is used in real time and is updated frequently with new information, the archive estimator can often give an answer after only a few trials that is as accurate as the frequency estimator can provide with 10 or even a 100 times as many trials.

The reason, of course, is that at a new updating of the network with new evidence, the archive estimator can access its memory of those trials which happen to conform to the new data.    For example, the new evidence instantiates a node that had not previously been observed. In this case, as long as the new evidence had some positive probability, some trials conforming to this new data are likely to already be in the archive. In other cases, the new evidence returns to a situation encountered at an earlier time. Of course, the bias and accuracy of the archive estimator is, initially, dependent on how the archive was originally constructed.    But after a search process  satisfying the condition above runs, sufficient probability mass is added to $P^T(\underline{X_E})$ to eliminate the initial bias.

A frequency estimator aggregates individual trials into the frequency tallies. This aggregation loses information as to which terms in the sum conform to a revision in the evidence. Thus when the network must be updated because of new evidence, the tallies must be reset to zero.

In settings where the prior probability of the evidence is small, there may be only a few trials conforming to the data that actually have positive probability. Often, it takes many trials to find the few that have positive probability. When the network is updated, the archive may already have 80% of the probability mass associated with the new evidence. This allows the archive based estimator to be quite accurate at the start, before any simulated trials are generated.

## 3. SIMULATION AS SEARCH.

A Monte Carlo simulation generates a set of independent and identically distributed trials. The new trial, if unique, increases the probability mass of the archive. Considered as a search technique, each attempt to generate a new trial ignores the trials that have come before it. Genetic search, on the other hand,  uses this history to obtain a picture of the explored region of  the sample space and determine which regions should be explored next.

A genetic algorithm[10]  is a search technique to find the maximum of a function over a set of possible solutions. The technique represents the various solutions as a binary string, called a genotype. There is a function, called the fit, which evaluates the proposed solution.  The method starts from an initial set of trial solutions (which may be generated via simulation) called the initial breeding population.  At each step, the algorithm selects parents from the breeding population representing the best genotypes generated in the previous generations via a random selection  method. The selection method is biased toward the individuals in the breeding population with greater fit (survival of the fittest), and creates children that have characteristics from both parents (crossover) with some deviation (mutation). The new generation is evaluated for the fit of the individuals. The process of creating a new generation and evaluating the fit of the individuals is repeated. With judicious choice of parameters that control the selection, crossover and mutation of the genotypes, as well as the size of the generation, the algorithm will come close to a global maximum.

In applying the techniques of genetic search to the problem of solving a Bayesian network, a trial solution of the network is represented by a string of bits. With networks of binary nodes, each position in the string represents a node in the network. For networks of nodes with more than 2 states, each node is represented with enough bits in the string to enumerate all the possible states of the node. The fit of the individual is naturally the joint probability of the trial solution.    The genetic algorithm will search for the individual with the greatest joint probability.  If there is evidence concerning the states of some of the nodes, either the search is constrained  to the trials that agree with these observations or the fit function penalizes those solutions that do not agree.

The genetic algorithm will search for the trial solution with the greatest fit (probability) that agrees with the evidence.  This trial solution is the  most probable explanation of the evidence. Along the road to finding this most likely explanation, the genetic algorithm will also discover many other solutions with significant probability. All unique trial solutions are added to the archive.

---

[10]  Goldberg [1989].   Rojas-Guzman & Kramer [1994] first applied genetic search to Bayesian networks for abductive reasoning (finding  the best solution that fits the evidence).

The probability mass of the trial solutions approximates the probability of the observed values. The Archive Estimation Theorem shows that, as the mass converges, the archive estimators approximate the solution to the network.

## 3.1 GENETIC SEARCH PARAMETERS

*Genetic search algorithms* have a terminology based on the theory of evolution. We define this terminology in the language of Bayesian networks.

**genotype.** In the genetic algorithm, there is a basic representation of the individual trial solutions. This is called the genotype and is a string of bits, where each bit represents an allele or characteristic. A solution in the context of the belief network is an assignment of states to each of the nodes. This assignment can be represented as a string of bits (see figure 3.1) and fits naturally into the genetic algorithm paradigm.

**Fit function.** The solution set is represented, thus, as a set of genotypes. Each genotype would, in the biological world, represent the genetic characteristics of an individual specimen. The characteristics enable the individual to adapt to the environment and achieve its goals. The fit of the individual is an evaluation of its characteristics toward meeting these goals. In a Bayesian network, the fit is the probability of the assignments of states to nodes indicated by the genotype.

The genetic algorithm is a optimization strategy adapted from notions of survival and evolution in biology. Given a current population, breeding results in evolutionary changes to the population that enable it to adapt to the environment and survive. The fitness (chance of survival) of the typical individual is thus being maximized.

In a genetic algorithm, one typically needs an **initial breeding set** to breed from. Such an initial set comes from the archive and is supplemented by a brief Monte Carlo simulation.

After the generation of the initial breeding population, a series of generations each consisting of many trial solutions are bred. The characteristics of the breeding cycle are:

1. **Survival of fittest**: The generation is bred from a breeding population consisting of the best trial solutions encountered so far.

2. **Random selection.** Parent pairs from the breeding population are randomly selected (with replacement). The probability of being selected is equal to the share of the fit value of the parent to the total fit of the set of breeders.

3. **Genotype crossover.** Once the parent pairs have been selected, the genotypes are combined to form offspring. This process is called crossover and amounts to splicing a group of characteristics of one parent with the remaining characteristics of the other parent. Essentially a subset of the bits in the genotype of one parent is replaced by the corresponding bits from the second.

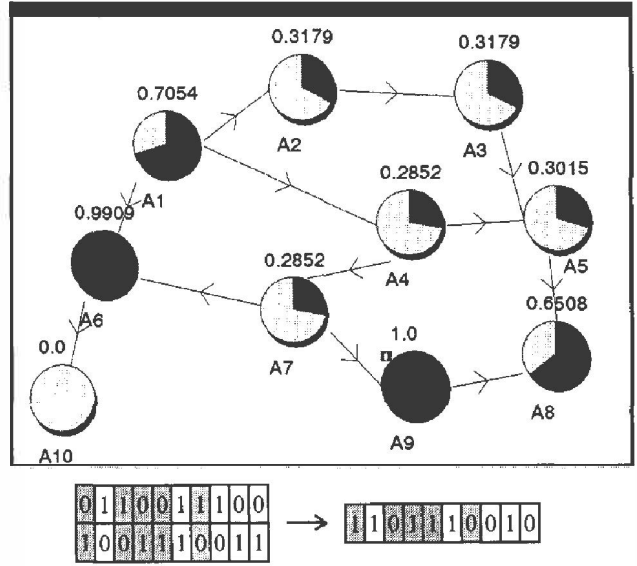

Figure 3.1   Genotype Crossover in a Bayesian network of 10 binary nodes. A4 is selected as the center of the Markov neighborhood of radius 1, {A1, A3, A4, A5, A7 } in the network. The lower diagram shows the genotype of two parents (left of the arrow) and an offspring after splicing the bits corresponding to the Markov neighborhood (shaded blocks) from the second parent into the corresponding bits of the first parent.

Both the decision to crossover and the set of bits to splice are determined at random. For the application to Bayesian networks, a node in the network is selected at random and a Markov neighborhood centered on the node of random diameter are selected. The bits in the genotype corresponding to the states of the nodes in the Markov neighborhood constitute the set of bits for splicing. See figure 3.1. A Markov neighborhood of radius 1 of a node $\alpha$ consists of the node, its parents, its children and the parents of its children:

$$M(\alpha) = \{\alpha\} \cup \xi(\alpha) \cup \xi(\xi^{-1}(\alpha)).$$

A Markov neighborhood of radius k is obtained by applying M k times, $M^k(\alpha) = M(M^{k-1}(\alpha))$, where $M^0(\alpha) = \alpha$.

The boundary of a Markov neighborhood isolates the interior of the neighborhood from the exterior or the rest of the network (the interior and exterior are conditionally independent given the boundary). Splicing genotypes on Markov neighborhoods naturally provides offspring that have compatible characteristics.

4. **Mutation.** The final step in determining the offspring is to randomly mutate the genes in the genotype. Each node that has not been instantiated (not in **E**) is subjected to mutation, independently. The probability distribution over the remaining possible states of the node is the conditional probability of the node (as specified by its link matrix) given the state assignments of the other nodes in the rest of the genotype. This reduces to the probability of the node given its Markov blanket.

Crossover and mutation keep the algorithm from settling on local maxima. They balance the need to explore a region that looks promising with the need to search unexplored regions.

**5. Evaluation and archiving.** Once the genotype of the offspring is determined, the genotype is added to the archive if it is a new trial. The offspring replaces a genotype in the breeding population if it has a better fit.

This process of selection, crossover, mutation, archiving, and evaluation is repeated until a new generation has been filled. This completes an evolution cycle. The search continues until either the archive conditional mass has not appreciably increased for a specified number of generations or the maximum cycle or time limit is exceeded.

The parameters that control the genetic search are the following (the values in parentheses are default values used in the comparisons of sections 4 and 5).

1. generation size (50)
2. breeding population size (40)
3. maximum number of generations
4. crossover probability (0.85)
5. mutation probability (0.01)

# 4. COMPARISON OF ALTERNATIVE METHODS FOR REAL TIME UPDATING: A SPECIFIC EXAMPLE.

The various methods of approximation, both those based on Monte Carlo simulation using frequency and archive estimators and those based on genetic search have been incorporated into **Bayes-On-Line™**, an application product built with Gensym's intelligent real time system, **G2®**.

The results from estimation of a specific sample Bayesian network are presented in this section. These results confirm these hypotheses:

1) Traditional Monte Carlo methods:

   * Forward sampling will provide a more efficient simulation than logic sampling, especially when the prior probability of the evidence is small.
   * When there is observed evidence, backward sampling often is more efficient, especially when the prior probability of the evidence is small.
   * However, several hundred trial simulations may be required before the advantage of backward sampling is realized.

2) Frequency vs. Archive estimators.

   * Frequency estimators will be more efficient when no evidence is presented to the network.

   * Archive estimators will prove more efficient when there is evidence and especially during sequential real time updating.

3) Genetic search is especially beneficial

   * when evidence has low prior probability or
   * when presented in a real time sequential updating environment.

## 4.0 THE SAMPLE NETWORK.

Figure 4.1 displays a sample network consisting of 32 binary nodes, including 7 logic (deterministic) gates. This network has a solution space of $2^{32} \cong 4.3$ billion trials and also has several undirected cycles. There are 7 leaf nodes or sensors labeled $S_1 ... S_7$.

The state of the network displayed in figure 4.1 is the exact solution when the sensor nodes S1, S4 and S5 are true and S3 is false. The root mean squared error of the estimated belief in each node to the belief in the exact solution is the goodness of fit measure used in these comparisons. Thus this measure is the average error in terms of probability of each estimated belief from its true value.

## 4.1    MONTE CARLO SAMPLING METHODS.

All three sampling methods, logic sampling, forward sampling, and backward sampling are identical when there is no evidence. Their performance is identical and is graphed in figure 4.3.

Table 4.1 displays the root mean squared error of the estimated solution to its exact value for both estimator types (frequency or archive) and the probability mass of the archive along with the probability mass of those trials in the archive that conform to the evidence.

Figure 4.2 displays a graph of the RMSE from logic, forward and backward sampling methods for the case of 4 simultaneously observed nodes ($S_1$ = T, $S_3$ = F, $S_4$ = T, $S_5$ = T).[11] The prior probability of $\underline{X}_E$ is approximately $6.5 * 10^{-4}$.

As expected, forward sampling clearly dominates logic sampling, although both are close after 10,000 trials. The graph shows that it took about 500 logic sampling trials before any trial was found that conforms to the evidence. The rmse for logic sampling remains quite high, above .2, for 7000 trials. Forward sampling forces conformity from the start and drops below 0.10 after 4500 trials.

Backward sampling also enforces conformity from the start. After a slow start, backward sampling nearly dominates the others, falling to below the 0.10 level after 2000 trials and achieving .033 after 10,000 trials. Note, however, that there is an initial phase (the first 1000 trials) where evidence based sampling is dominated by forward sampling.

---

[11] In our earlier notation, $\mathbf{E}$ = { $S_1$, $S_3$, $S_4$, $S_5$} and $\underline{X}_E$ = {T, F, T, T}.

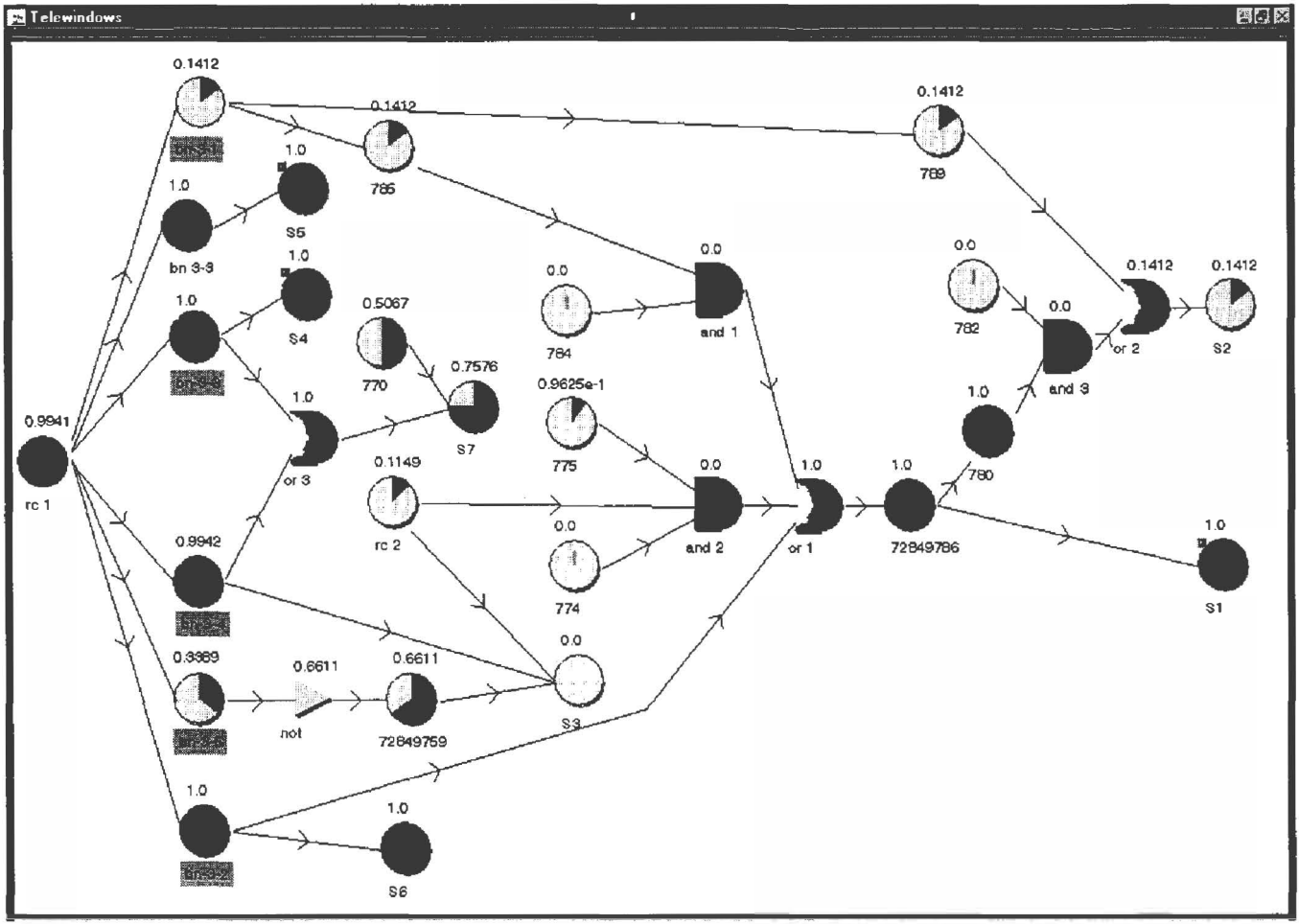

Figure 4.1. The sample network.

TABLE 4.1  Comparison of simulation strategies:
Network in figure 4.1

| experiment | #obs | update | simulation | | genetic search | | | RMSE | | probability masss | |
|---|---|---|---|---|---|---|---|---|---|---|---|
| | | | method | trials | #gen | gen size | archive | frequency | archive | cond |
| Fwd-0-obs | 0 | | fwd | 10000 | | | 0.005081 | 0.005486 | 0.9842 | 0.9842 |
| GA/fwd-0-obs | 0 | | fwd | 5000 | 50 | 50 | 0.000090 | 0.007018 | 0.9701 | 0.9701 |
| Fwd-4-seq | 4 | seq | fwd | 2000 / obs | | | 0.109000 | 0.130100 | 0.9468 | 0.0004 |
| Bwd -4-seq | 4 | seq | bwd | 2000 /obs | | | 0.110300 | 0.180400 | 0.9416 | 0.0004 |
| GA/fwd-4-seq | 4 | seq | fwd | 1000/obs | 20 | 50 | 0.006286 | 0.158800 | 0.9170 | 0.0005 |
| GA/bwd-4-seq | 4 | seq | bwd | 1000/obs | 20 | 50 | 0.000091 | 0.141700 | 0.9099 | 0.0006 |
| Fwd-4-all | 4 | all | fwd | 10000 | | | 0.1067 | o.04267 | 0.00043 | 0.00042 |
| Bwd-4-all | 4 | all | bwd | 10000 | | | 0.02643 | 0.03279 | 0.00061 | 0.00061 |
| GA/fwd-4-all | 4 | all | fwd | 5000 | 50 | 50 | 0.001094 | 0.072360 | 0.0014 | 0.0006 |
| GA/bwd-4-all | 4 | all | bwd | 5000 | 50 | 50 | 0.000033 | 0.051770 | 0.0014 | 0.0007 |

It should be noted that the computational cost of generating a new trial is the same between logic and forward sampling, while backward sampling is somewhat higher, since several Bayes inversions are computed for each trial.

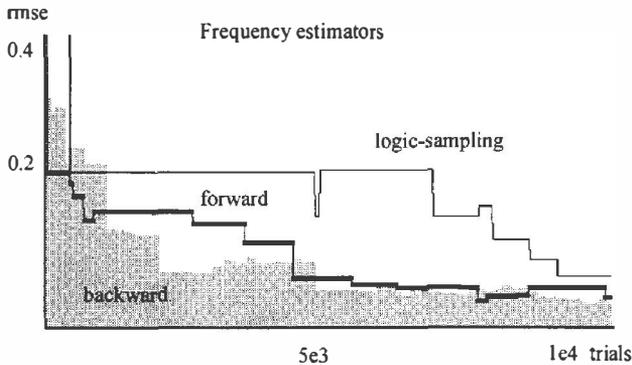

Figure 4.2. Comparison of the rmse for the frequency estimator when logic, forward and backward sampling are applied to the network of fig 4.1 with 4 observations introduced simultaneously.

## 4.2 FREQUENCY VS. ARCHIVE ESTIMATORS.

The graph in Figure 4.3 shows that the frequency estimator dominates the archive estimator when there is no data presented to the network and a Monte Carlo sampling method is used to generate new trials. The graph of the archive estimator is monotone decreasing (no forgetting of older trials). The statistics in Table 4.1 show that the probability mass of the unique solutions that are discovered after 10,000 simulations have 98.4 % of the maximum mass possible. 90% of the probability mass is found in the first 1000 trials. The rmse achieved is 0.005.

Figure 4.4 shows that when there is evidence presented to the network and the evidence based sampling method is used in a Monte Carlo simulation, the archive estimator closely tracks, and somewhat dominates the frequency based estimator, achieving a 1/3 reduction in the rmse (.0264 vs. .033). The probability mass attains 0.0006 which is close to the prior probability of the evidence. Note that the graph in Figure 4.5 indicates that the archive estimators do not show an improvement when either logic or forward sampling methods are used (compare to Fig 4.2).

The data in table 4.1 and figure 4.6 shows the effect of applying genetic search, utilizing evidence based simulation to establish the initial set of breeders. The experiment will first run 5000 trial simulations and then start genetic search for 50 generations with 50 individuals bred in each generation. The initial breeding set consists of the best 40 trials in the archive after the simulation phase. The vertical axis marks the end of the simulation phase.

Almost immediately the genetic algorithm finds most of the remaining solutions that are consistent with the evidence and have positive probability. As a result of genetic search, the archive rmse falls from 0.039 to 0.000033. The conditional probability mass achieved is 0.00065 which is a 7% increase over the value achieved after 10,000 trials with evidence based simulation (table 4.1).

Generating a new individual with a genetic algorithm requires computational resources on the order of magnitude of a forward sampled trial and less computational resources than generating a new trial with evidence based simulation. A run of 10,000 trials of evidence based simulation required computation time 6 times greater than a combination of 5000 backward simulations and 2500 genetic search trials. Even in that case, the archive rmse was 0.000042, much better than the lengthy simulation.

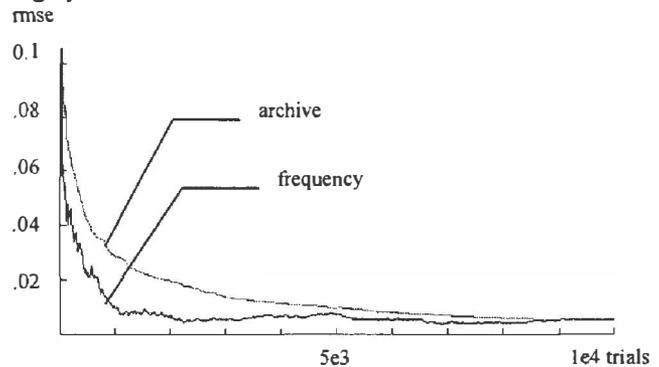

Figure 4.3. RMSE of archive and frequency, forward simulation with no observed nodes.

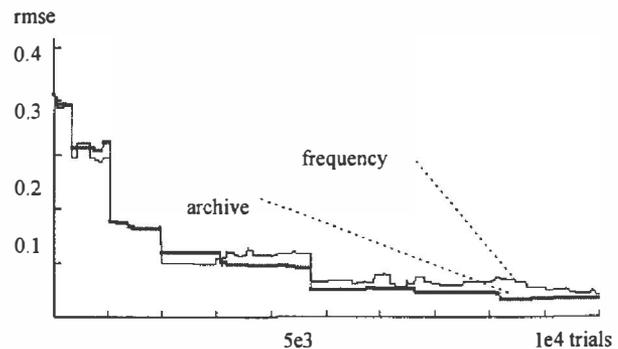

Figure 4.4. RMSE from backward sampling using the frequency estimator of Monte Carlo simulation and the archive estimator. In this case the network is updated introducing 4 observed nodes simultaneously.

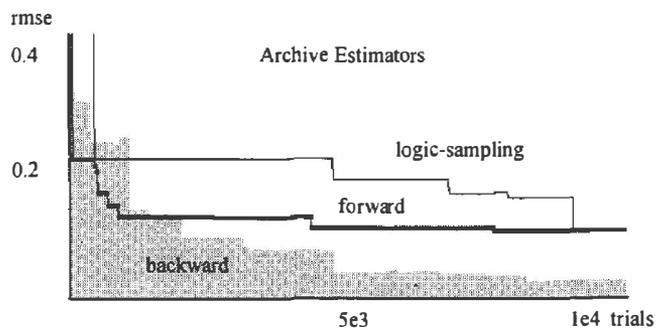

Figure 4.5. The rmse of the archive estimator using logic, forward and backward sampling methods in simulation of the network with 4 observed nodes introduced simultaneously.

## 4.3 SEQUENTIAL UPDATING.

When a Bayesian network is employed on-line in a real time environment, evidence is invariably introduced sequentially. Especially when new evidence instantiates a previously unobserved node, the trial archive will have some trials that conform to the new evidence and an archive based estimator will be immediately available. A smaller number of simulation trials will be required to achieve the accuracy desired. An estimation strategy that utilizes what is already in the archive, a short period of backward simulation to enhance the initial breeding set and a few rounds of genetic search can provide a sufficiently accurate solution within the time frame demanded by the application.

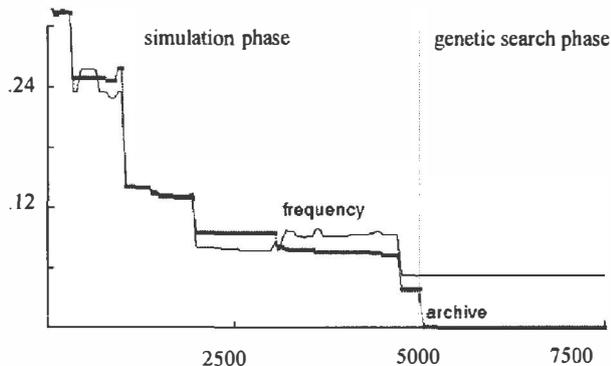

Figure 4.6. The results from introducing genetic search. The network is simulated using backward sampling for 5000 trials. The best 40 trials forms an initial breeding population. Genetic search is then applied for 50 generations of 50 trials per generation. The rmse of frequency and archive estimators are plotted in the diagram.

The first experiment (figure 4.7) splits up the 10,000 trials into 5 simulations of 2000 trials each. The 5 simulations introduce the evidence sequentially: no evidence, S1 is set to true, S3 is set to false, S4 to true and finally S5 to true. In figure 4.7, the five regions clearly mark the introduction of new evidence. Since forward sampling performs poorly with lower probability of the observed evidence, both frequency and archive estimators provide progressively higher rmse's. In the first panel corresponding to no evidence, the frequency estimator dominates, but in each of the remaining cases, the archive estimator dominates. In each of the latter cases, the archive estimator benefits at the start of the simulation from the existence of trials conforming to the new evidence in the archive.

The backward sampling method has the same qualitative result (figure 4.8): As more evidence is introduced, the rmse increases, indicating a longer simulation is required to achieve a given accuracy level as each new piece of evidence is introduced. In comparing figure 4.8 with figure 4.7, one cannot help but notice the poor performance of the frequency estimator with backward sampling as opposed to forward sampling. This is consistent with figure 4.5. Evidence based sampling only dominates forward sampling after the initial 2000 trials. Nevertheless, figure 4.8 clearly shows the superiority of the archive estimator when using backward sampling, as noted in section 4.2 above.

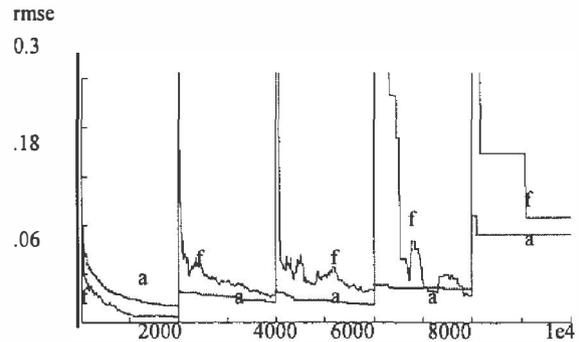

Figure 4.7. Rmse from simulation of sequential updating. Shown are the errors from frequency (f) and archive (a) estimators using the forward sampling method.

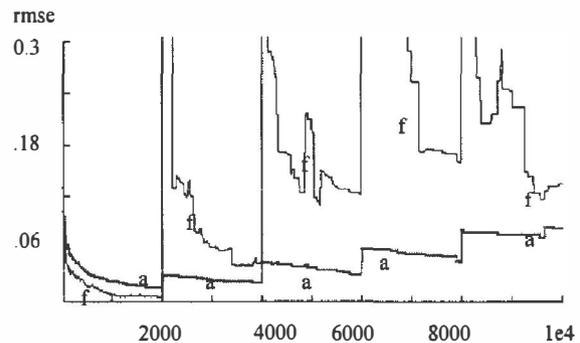

Figure 4.8. Rmse from simulation of sequential updating. Shown are the errors from frequency (F) and archive (A) estimators using the backward sampling method.

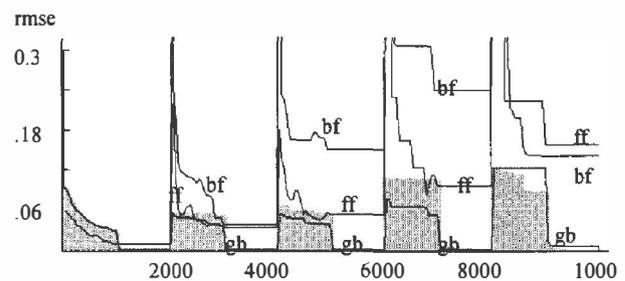

Figure 4.9. Sequential updating using genetic search with an initial phase of simulation (1000 trials) to generate the breeding set. Results from using either forward or backward during the simulation phase are shown. The shaded region is the rmse from using genetic search an with initial phase of forward simulation. gb indicates with an initial phase of backward simulation. The rmse of the frequency estimator with initial phases from forward or backward are indicated by ff and bf.

In the final experiment, 10,000 trials are generated, 2000 with the introduction of each evidence (figure 4.9). After the introduction of a new piece of evidence, the backward simulation method generates 1000 trials. Genetic search is then started using the best 40 trials from the archive as the initial breeding population. The genetic algorithm then runs until another 1000 trials have been generated. Each time the rmse of the archive estimator drops precipitously. At the end of the updating sequence, the rmse achieved is comparable (in the same order of magnitude) to the result observed earlier when the evidence is introduced all at once (figure 4.6).

# 5. GENERALITY OF THE RESULT.

To investigate the generality of the comparisons of approximation methods, the same experiments were conducted on randomly generated Bayesian networks. A similar pattern was found. Especially in situations where the prior probability of the evidence is very small and where the evidence is introduced sequentially, the combination of an initial period of simulation followed by a genetic search provides a profound improvement over forward and backward sampling techniques.

The random networks were all constrained to have 32 nodes. The number of possible states for each node was selected at random between 2 and 4 with a bias toward the smaller number of states. The number of parents for each node was selected at random (between 0 and 3). The selection of the parents was also randomized in a manner that guarantees that the network is a directed acyclic graph. The entries in the conditional link matrices of each node were selected in the following manner that gives a bias toward having many 0 entries. Each cell is given a 0.5 probability that it has a value of 0.0.[12] If it is to have a positive value, its value is selected uniformly between 0 and 1. These choices are constrained so that each row of the matrix sums to 1.0. A set of four leaf nodes with the lowest prior probability was selected as the evidence set for each network.

The average of RMSE and probability mass using both the archive and frequency estimators are given in Table 5.2. These randomly generated networks are somewhat more difficult to estimate with any of the simulation methods than the example of section 4. This is reflected in the fact that 10000 trials was not sufficient to bring the probability mass of the archive above 0.5

Again we see that the frequency estimator of a forward simulation is superior when no nodes have been observed (fwd-0-obs). Although less than half of the total probability mass has been generated, the average RMSE over all 12 networks is below 1%. Since there is still considerable unexplored probability mass, it will take considerably more trials before the archive estimator will catch up to the frequency estimator.

---

[12] The bias toward 0 entries reflects the prior belief that the link matrices will be sparse in practice (e.g. deterministic nodes or noisy gates -- such as the noisy or-gate model).

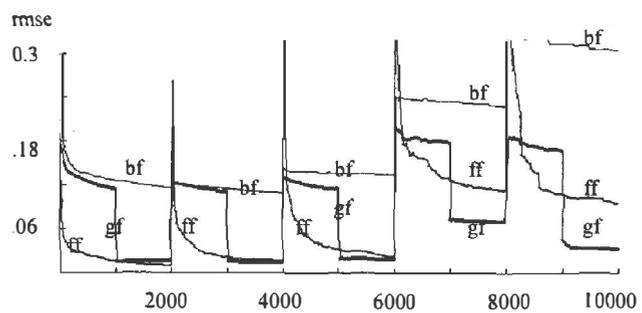

rmse

Figure 5.1. Rmse from simulation of sequential updating. ff = forward sampling with frequency estimator. bf = backward sampling with archive estimator. gf = genetic search with initial forward simulation for generating a breeding set, archive estimator.

With all 4 observations introduced simultaneously (fwd-4-all) 10000 forward simulations achieves a 6% error rate when the frequency estimator is used. Backward simulation (bwd-4-all) does not show much promise. As we had observed in section 4, the archive estimator is better than the frequency estimator with backward simulation. But a 28% error rate is unacceptable. One conjecture is that 10000 trials remains in the initial region where forward simulation dominates backward simulation (figure 4.2) in these networks. Combining 5000 forward simulation trials followed by genetic search (GA/fwd-4-all) reduces the error to under 3% using the archive estimator.

When the evidence must be introduced sequentially, the combination of forward simulation and genetic search results in a three times more accurate estimate than pure forward simulation. The graph in figure 5.1 illustrates the precipitous reduction in the average RMSE each time the genetic search algorithm is applied after an initial round of forward simulation. See also table 5.2. Table 5.1 provides the comparison between forward simulation frequency estimator and the combination of forward simulation and genetic search with the archive estimator at the end of each observation phase.

TABLE 5.1 Comparison of rmse from forward and forward/GA methods during sequential updating.

| No Observations | FWD-4-seq* | GA/fwd-4-seq** |
|---|---|---|
| 0 | 0.0091 | 0.0173 |
| 1 | 0.0161 | 0.0149 |
| 2 | 0.0221 | 0.0193 |
| 3 | 0.1134 | 0.0714 |
| 4 | 0.0951 | 0.0337 |

\* Frequency estimators

\*\* Archive estimators

TABLE 5.2 Comparison of simulation strategies:
Average of 12 randomly generated networks

| experiment | #obs | update | simulation | | genetic search | | RMSE | | probability mass | |
|---|---|---|---|---|---|---|---|---|---|---|
| | | | method | trials | #gen | gen size | arch | frequency | arch | cond |
| Fwd-0-obs | 0 | | fwd | 10000 | | | 0.090430 | 0.004200 | 0.4277 | 0.4277 |
| GA/fwd-0-obs* | 0 | | fwd | 5000 | 100 | 50 | 0.011700 | 0.004200 | 0.4305 | 0.4305 |
| Fwd-4-seq | 4 | seq | fwd | 2000 / obs | | | 0.148300 | 0.095100 | 0.2869 | 0.0018 |
| Bwd-4-seq | 4 | seq | back | 2000 /obs | | | 0.304700 | 0.671400 | 0.2646 | 0.0012 |
| GA/fwd-4-seq* | 4 | seq | fwd | 1000 /obs | 20 /obs | 50 | 0.033700 | 0.104100 | 0.3058 | 0.0019 |
| Fwd -4-all | 4 | all | fwd | 10000 | | | 0.134600 | 0.066800 | 0.4284 | 0.0020 |
| Bwd-4-all | 4 | all | back | 10000 | 100 | 50 | 0.280600 | 0.449600 | 0.4277 | 0.0014 |
| GA/fwd-4-all* | 4 | all | fwd | 5000 | | | 0.026600 | 0.072400 | 0.4320 | 0.0021 |

* All genetic search used a breeding size of 40,

0.85 crossover probability and 0.01 probability of mutation

## 6. CONCLUDING REMARKS.

The results demonstrated in sections 4 & 5 provide the basis for efficient real time updating of a Bayesian network when time only permits an approximate solution. Utilize a Monte Carlo sampling method (e.g., forward or backward sampling) to generate a sufficient number of trials for an initial breeding set. Then run a genetic search procedure until either the time allocated for a real time response runs out, the accuracy reaches a tolerance level, or the graph of the archive probability mass from trials that conform to the evidence is flat. Use an estimate based on partial sums of the joint probabilities of trials in the archive rather than on the weighted frequency of the values encountered during simulation.

While it would appear that maintaining an archive of the trials generated would be very costly, in fact modern computers are quite capable of accomplishing the task. An id code for each trial is simply a binary string representing the values of each node interpreted as a vector of integers. Searching for a match between two vectors of integers involves very primitive computer operations. Furthermore, as the prior probability of the evidence diminishes, the number of items in the archive conforming to the evidence will in fact be small. Hence the test of whether a trial is already in the archive is not that expensive.

A study of the complexity of the genetic search algorithm vs. forward simulation has not been attempted. However, the experience with the random networks encountered in this paper suggests that the computational cost of a cycle of 100 forward simulations is similar to breeding 100 members of a new generation. Use of backward sampling during the simulation phase is not recommended. For one, generating a new trial via backward sampling is quite expensive, especially if the number of ancestors of the evidence set is large. In addition, there appears to be an initial phase during which forward simulation dominates backward simulation (figure 4.2 and 4.5). Although for a special case (like figure 4.1 and table 4.1), backward sampling paid off, this was not the situation for the more general study of section 5.

There are a number of parameters in this strategy to be tuned for each application. These include the parameters of the genetic search and the mix between the number of initial simulations for generating an initial breeding population. We have kept these parameter values constant through all the experiments and all random networks. The use of the genetic search shows a precipitous drop in rmse and there appears to be little gain from additional search with these parameter values. It would appear that further research into this mix of simulation and genetic search and the tuning of these parameters would be fruitful.

## 7. RELATED LITERATURE.

Druzdzel [1994] has characterized the nature of the joint probability distribution $P(X_N)$ of typical Bayesian networks. Given that the solution space is badly chopped up with the probability mass concentrated in a small, but broken and disconnected, region, it should come as no surprise that genetic search methods prove to be beneficial.

Santos and Shimony [1994] devised a non-random search method for enumerating the trials in the solution space that have the most probability mass. Their method is based on solving a fairly complex linear programming problem. In their concluding remarks, they also suggested that the approach that has been taken here could prove to be fruitful. Not only does it appear that their conjecture is correct, but there is also considerable gain in simplicity of the algorithm over the ILP based approximation method.

## 8. APPENDIX

### 8.1 FORMAL DEFINITION OF BACKWARD SIMULATION:

The following steps define the formal definition of this sampling distribution:

Let $A(\beta)$ denote the ancestors of $\beta$: $\alpha \in A(\beta)$ if and only if there exists a $k \geq 0$ such that $\alpha \in \xi^k(\beta)$, where $\xi^k(\beta) = \xi(\xi^{k-1}(\beta))$, $\xi^0(\beta) = \beta$.

Let $A(E)$ denote ancestors of the evidence nodes in $E$,
$$A(E) = \cup_{\beta \in E} A(\beta). \text{ Note that } E \subset A(E).$$

Let $\{\alpha_1 ... \alpha_n\}$ be an ordering of the nodes so that any two nodes $a_i$ and $a_k$ satisfy:

i < k if either $\alpha_i \in E$ and $\alpha_k \notin E$,

or $\alpha_i \in A(E)$, $\alpha_k \in A(E)$ and $\alpha_k \in \xi(\alpha_i)$,

or $\alpha_i \in A(E)$ and $\alpha_k \notin A(E)$,

or $\alpha_i \notin A(E)$, $\alpha_k \notin A(E)$ and $\alpha_i \in \xi(\alpha_k)$

Clear the set $B = \varnothing$.

Following the node ordering, assign values to each node $\alpha_i$ as follows.

i) if $\alpha_i$ is an evidence node then $X_{\alpha_i} = \underline{X}_{\alpha_i}$.

ii) if $\alpha_i$ has already been assigned then go on to $\alpha_{i+1}$

iii) if $\alpha_i \in A(E) \backslash E$ , then :

Let $\beta$ be a child of $\alpha_i$.

Define $\zeta^1(\beta) = \xi(\beta) \cap \{\alpha_1...\alpha_{i-1}\}$, to be the parents of $\beta$ that are earlier in the ordering than $\alpha_i$ and

Define $\zeta^2(\beta)$ to consist of $\alpha_i$ and the remaining parents of $\beta$,
$\zeta^2(\beta) = \xi(\beta) \cap \{\alpha_i...\alpha_n\}$

By the requirements of the node ordering, $\beta$ and the parents in $\zeta^1(\beta)$ have already been assigned a value.

Sample $\zeta^2(\beta)$ simultaneously from the Bayes inverse of the link matrix:

$P(X_{\zeta^2(\beta)} \mid X_\beta, X_{\zeta^1(\beta)}) = \kappa_\beta P(X_\beta \mid X_{\zeta^1(\beta)}, X_{\zeta^2(\beta)})$
where $\kappa_\beta$ satisfies

$1/\kappa_\beta = \Sigma X_{\zeta^2(\beta)} P(X_\beta \mid X_{\zeta^1(\beta)}, X_{\zeta^2(\beta)})$.

Insert $\beta$ into the set of sampled nodes $B$.

iv) For the remaining nodes, $\alpha_i \notin A(E)$, $\xi(\alpha_i) \subset \{\alpha_1...\alpha_{i-1}\}$ -- the parents of $\alpha_i$ have already been assigned. $\alpha_i$ can be simulated from the link matrix $P(X_{\alpha_i} \mid X_{\xi(\alpha_i)})$.

The sampling distribution determined in this manner is:

$$p^S(X_N) = \Pi_{\beta \in B} \kappa_\beta P(X_\beta \mid X_{\zeta^1(\beta)}, X_{\zeta^2(\beta)}) \bullet$$
$$\Pi_{\alpha \notin A(E)} P(X_\beta \mid X_{\xi(\alpha)})$$

The likelihood weight is:

$$z(X_N^t) = \frac{\Pi_{\alpha \in A(E)} P(X^t_\alpha \mid X^t_{\xi(\alpha)})}{\Pi_{\beta \in B} \kappa^t_\beta P(X_\beta^t \mid X^t_{\zeta^1(\beta)}, X^t_{\zeta^2(\beta)})}$$